\documentclass[letterpaper,twocolumn,10pt]{article}
\usepackage{usenix-2020-09} 
\usepackage{graphicx}
\usepackage{hyperref}
\usepackage{booktabs}
\usepackage{enumitem}
\usepackage{listings}
\usepackage{algorithm}
\usepackage{algorithmic}
\usepackage{float}
\usepackage{amsmath}
\usepackage{caption}

\title{\Large \bf Fault-Tolerant Sandboxing for AI Coding Agents: A Transactional Approach to Safe Autonomous Execution}

\author{
{\rm Boyang Yan}\\
University of Virginia\\
\texttt{rhe9cf@virginia.edu}
}

\begin{document}
\maketitle

\begin{abstract}
The transition of Large Language Models (LLMs) from passive code generators to autonomous agents introduces significant safety risks, specifically regarding destructive commands and inconsistent system states.
Existing commercial solutions often prioritize interactive user safety, enforcing authentication barriers that break the headless loops required for true autonomy.
This paper presents a Fault-Tolerant Sandboxing framework designed to mitigate these risks through a policy-based interception layer and a transactional filesystem snapshot mechanism.
We hypothesize that wrapping agent actions in atomic transactions can guarantee safety with acceptable latency, outperforming the heavy initialization overhead of containers or the interactive friction of commercial CLIs.
We validated this approach by deploying the Minimind-MoE LLM served via nano-vllm on a custom Proxmox-based testbed utilizing EVPN/VXLAN isolation.
Experimental results demonstrate a 100\% interception rate for high-risk commands and a 100\% success rate in rolling back failed states.
Crucially, our prototype incurs only a 14.5\% performance overhead (approx. 1.8s) per transaction.
In contrast, benchmarking against the Gemini CLI sandbox revealed that it requires interactive authentication ("Sign in"), rendering it unusable for headless, autonomous agent workflows.
\end{abstract}

\section{Introduction}
The increasing sophistication of Large Language Models (LLMs) has enabled AI coding agents to move beyond simple code generation into fully autonomous execution environments \cite{wang2023survey, li2023camel}.
Tools such as OpenInterpreter \cite{openinterpreter2023} and AutoGen \cite{wu2023autogen} allow LLMs to execute shell commands directly.
While this drives productivity, it elevates the risk profile significantly. An LLM agent effectively operates as a "black box" operator; it may generate commands that are syntactically correct but logically flawed, resulting in unintended consequences ranging from the deletion of critical system files (e.g., \texttt{rm -rf /}) to the introduction of security vulnerabilities.

However, a parallel trend in AI research—the shift towards Small Language Models (SLMs)—offers a new avenue for mitigating these risks. While massive LLMs (100B+ parameters) dominate general-purpose chat benchmarks, they are often prohibitively expensive and slow for the high-frequency, iterative loops required by autonomous agents. This work explores the intersection of \textit{transactional safety} and \textit{efficient inference}, leveraging SLMs to build agents that are not only safe but also economically viable for edge deployment.

The core problem addressed in this work is two-fold: (i) \textbf{Unsafe Execution}, characterized by a lack of real-time, policy-based validation; and (ii) \textbf{Lack of Fault Recovery}, characterized by the absence of a mechanism to restore a consistent state when a tool-call fails.

\subsection{Hypothesis and Contribution}
Our main hypothesis is that \textit{transactional fault-tolerance} can be implemented for AI coding agents to ensure state consistency and safety, provided a measurable, non-trivial performance penalty is accepted.
This work positions itself in the design spectrum between lightweight, unsafe local execution (e.g., Python \texttt{subprocess}) and heavyweight, persistent isolation (e.g., full VM provisioning).
We propose a "middle-ground" solution inspired by database transactions \cite{gray1981transaction}: treating every agent tool-call as an atomic operation (ACID).

The contributions of this paper are:
\begin{itemize}
    \item A formal definition of transactional semantics for LLM tool-use.
    \item A comprehensive analysis of the SLM vs. LLM trade-offs in the context of autonomous agents.
    \item A prototype implementation of a pre-execution snapshot mechanism.
    \item An evaluation on a realistic Proxmox/EVPN testbed using the Minimind-MoE model.
    \item A comparative analysis demonstrating why commercial tools (Gemini CLI) fail for headless autonomy.
\end{itemize}

The artifacts for this project are available in the GitHub repository \url{https://github.com/yanboyang713/Sandboxing-for-AI-Coding-Agents}.

\section{Background: The Case for Small Language Models}
The prevailing dogma in recent AI development has been "bigger is better," leading to models exceeding trillions of parameters. However, for autonomous agents operating in production environments, this scaling law hits diminishing returns. This section details the technical divergence between Large Language Models (LLMs) and Small Language Models (SLMs), motivating our choice of the latter.

\subsection{Defining the Divide: LLMs vs. SLMs}
The distinction between LLMs and SLMs is primarily defined by parameter count and training objectives. 
LLMs, such as GPT-4 or Llama 3 (70B+), are general-purpose "knowledge engines." They are trained on massive, indiscriminate datasets to maximize broad generalization capabilities \cite{weka_slm_llm}. 
In contrast, SLMs typically range from 1 million to 10 billion parameters \cite{huggingface_slm}. They are often trained or fine-tuned on highly curated, domain-specific datasets.
While LLMs excel at complex reasoning and creative generation, SLMs are designed for efficiency and specific task execution, such as code generation or network configuration.

\subsection{Architectural Efficiencies}
The efficiency of SLMs is not merely a product of fewer parameters but also of specialized architectural choices:
\begin{itemize}
    \item \textbf{Knowledge Distillation:} Many SLMs are trained using "teacher-student" frameworks, where a massive teacher LLM generates synthetic training data to transfer its reasoning capabilities to a smaller student model \cite{datacamp_slm}. This allows SLMs to achieve high performance on specific tasks without the massive computational overhead of the teacher.
    \item \textbf{Quantization:} SLMs are frequently deployed with aggressive quantization (e.g., 4-bit or 8-bit integers) rather than standard 16-bit or 32-bit floating-point numbers. This reduces memory footprint and increases inference speed on consumer-grade hardware \cite{prem_edge}.
\end{itemize}

\subsection{Why Agents Need SLMs}
For an autonomous coding agent, which may execute hundreds of "think-act" loops to solve a single problem, the overhead of LLMs becomes untenable. SLMs offer three critical advantages:

\textbf{1. Latency and The Interactive Loop:}
Autonomous agents operate in a loop: Observe $\rightarrow$ Think $\rightarrow$ Act.
Cloud-based LLMs introduce variable network latency and queue times, often taking seconds to generate a response. SLMs running locally can achieve sub-second inference times \cite{augment_slm}. This reduction in latency is critical for maintaining the "flow" of an agent, preventing timeouts in time-sensitive operations like network handshakes.

\textbf{2. Privacy and Data Sovereignty:}
Sending proprietary codebases or sensitive network configurations to a public LLM API poses significant security risks \cite{cloverdx_privacy}. SLMs allow for "air-gapped" intelligence. By running the model entirely within the local infrastructure (on-premise), organizations ensure that sensitive data never leaves their control plane. This is particularly vital for our testbed environment, which simulates critical infrastructure.

\textbf{3. Economic Viability:}
The cost of inference scales linearly with the number of active parameters. Running a 7B parameter SLM can cost 10--30$\times$ less per token than a 70B LLM \cite{hexaware_slm}. For an agent that might generate thousands of lines of logs and code trials, the economic argument for SLMs is decisive.

\subsection{The Role of Mixture of Experts (MoE)}
To bridge the gap between the reasoning power of LLMs and the efficiency of SLMs, we utilize the Mixture of Experts (MoE) architecture. 
As detailed in \cite{epoch_moe}, MoE models decouple model capacity (total parameters) from inference cost (active parameters). 
A gating network routes each token to a specific subset of "experts" (small neural networks). 
This means a model like Minimind-MoE can have a high total parameter count (high capacity for knowledge) but only activate a fraction of them for any given inference step. This sparse activation pattern is key to enabling high-intelligence agents on resource-constrained edge hardware.

\subsection{The Limitations of SLMs}
Despite their efficiency, SLMs introduce distinct trade-offs that must be managed. The primary drawback is a reduction in \textbf{reasoning depth} and \textbf{knowledge breadth} compared to their larger counterparts \cite{weka_slm_llm}. SLMs are more prone to "hallucination" when faced with tasks outside their specific training distribution. Furthermore, because they possess fewer parameters to store world knowledge, they may struggle with complex, multi-step planning tasks that require integrating disparate concepts \cite{huggingface_slm}.
Additionally, SLMs often have smaller context windows (the amount of text they can process at once). This limits their ability to ingest large codebases or extensive log files in a single pass, necessitating chunking strategies that can fragment context \cite{augment_slm}. These limitations suggest that while SLMs are ideal for the \textit{execution} of well-defined tasks, they may still require oversight or "hand-off" mechanisms for high-level architectural planning.

\section{Related Work}
The challenge of sandboxing untrusted code is not new, but the context of \textit{autonomous agents} introduces unique requirements for state recovery and headless operation.

\subsection{Traditional Sandboxing and Virtualization}
Standard containerization (e.g., Docker \cite{merkel2014docker}) provides excellent isolation by leveraging kernel namespaces and cgroups. However, managing ephemeral containers for every single agent command can introduce significant orchestration overhead. Furthermore, standard Docker containers do not inherently support rollback to a specific previous state within a running session without complex volume management or external snapshotting tools. Similarly, full virtual machine (VM) provisioning offers strong security guarantees but incurs startup latencies (often tens of seconds) that break the interactive think-act-observe loop required by autonomous agents.

\subsection{Language-Specific Virtual Environments}
Python virtual environments (e.g., \texttt{venv}, Conda) are the standard for dependency isolation in development. While they effectively isolate Python packages, they offer no protection against filesystem damage outside the library path. Moreover, they lack atomic rollback capabilities; if a \texttt{pip install} command fails halfway through, the environment is left in a corrupted, intermediate state that often requires manual cleanup.

\subsection{Commercial Agent Runtimes}
Emerging industry tools, such as the Gemini CLI \cite{gemini2024}, have begun incorporating checkpointing and command restrictions. However, a critical divergence in design philosophy exists. As our evaluation demonstrates, tools like Gemini CLI prioritize \textit{interactive safety} over \textit{headless automation}. The requirement for interactive authentication creates significant barriers for autonomous agent integration, rendering them less suitable for the high-frequency, unsupervised loops targeted by our work.

\section{System Design}
Our Sandbox framework imposes two layers of safety: a \textbf{Tool-Call Sandboxing Layer} for pre-execution validation, and a \textbf{Fault Recovery Framework} utilizing transactional filesystem checkpoints.

\subsection{Formal Model of Transactional Execution}
We define the agent's environment state at time $t$ as $S_t$.
A tool-call $C$ is a function mapping $S_t \rightarrow S_{t+1}$.
The validity of a command is determined by a policy function $P(C) \in \{Safe, Unsafe, Uncertain\}$.
We introduce a Transactional Wrapper $T(C)$ such that:
\begin{equation}
    S_{t+1} = 
    \begin{cases} 
      S_t + \Delta_{C} & \text{if } execution(C) \text{ succeeds} \\
      S_t & \text{if } execution(C) \text{ fails} 
   \end{cases}
\end{equation}
This ensures atomicity: the state is either successfully advanced or perfectly preserved.

\subsection{Command Interception Policy}
A lightweight policy engine parses incoming commands and classifies them:
\begin{enumerate}
    \item \textbf{Safe/Whitelisted:} Read-only or low-risk commands (e.g., \texttt{git status}, \texttt{ls}). These bypass the snapshot mechanism to reduce latency.
    \item \textbf{Unsafe/Blacklisted:} Destructive commands (e.g., \texttt{rm -rf /}, \texttt{mkfs}). These are immediately blocked.
    \item \textbf{Uncertain/Requires Checkpoint:} State-modifying commands (e.g., \texttt{pip install}, \texttt{sed -i}). These trigger the transactional recovery flow.
\end{enumerate}

\subsection{The Snapshot-Rollback Algorithm}
The core logic of the fault recovery system is detailed in Algorithm 1. The system utilizes a copy-on-write simulation using \texttt{shutil} to create a restore point before any "Uncertain" command is executed.

\begin{algorithm}[H]
\caption{Transactional Execution Loop}
\begin{algorithmic}[1]
\REQUIRE Command $C$, Current State $S_{curr}$
\STATE Classify $C$ using Policy Engine $P(C)$
\IF{$P(C) = \text{Unsafe}$}
    \RETURN \textsc{Error}("Policy Violation")
\ENDIF
\IF{$P(C) = \text{Safe}$}
    \STATE Execute $C$
    \RETURN Output
\ENDIF

\STATE \textbf{Prepare Phase:}
\STATE $S_{snap} \leftarrow \textsc{Snapshot}(S_{curr})$
\STATE Output $\leftarrow$ \textsc{Execute}(C)
\IF{ExitCode $\neq 0$}
    \STATE \textbf{Rollback Phase:}
    \STATE Restore $S_{curr} \leftarrow S_{snap}$
    \RETURN \textsc{Error}("State Rolled Back")
\ELSE
    \STATE \textbf{Commit Phase:}
    \STATE Discard $S_{snap}$
    \RETURN Output
\ENDIF
\end{algorithmic}
\end{algorithm}

\section{Implementation}
To validate the hypothesis, we implemented the prototype in Python. The complexity of the implementation lies not just in the software logic, but in the underlying infrastructure that supports the agent.

\subsection{Testbed Infrastructure: Proxmox \& EVPN}
Experiments were conducted on a custom \href{https://yanboyang.com/posts/2025-04-10-010004-data_center_testbed_design}{Data Center Testbed} designed to simulate a realistic production cloud environment.
\begin{itemize}
    \item \textbf{Virtualization Layer:} The cluster runs on \textbf{Proxmox VE 9.0} \cite{proxmox}. The agent operates within a dedicated LXC container, while the Inference Server runs on a separate VM with GPU passthrough.
    \item \textbf{Network Isolation (EVPN/VXLAN):} Unlike simple VLANs, we utilized Ethernet VPN (EVPN) \cite{rfc7432} with VXLAN encapsulation \cite{rfc7348}. The network core is powered by \textbf{VyOS} routers acting as VTEPs (VXLAN Tunnel Endpoints). This creates a Layer 3 leaf-spine topology, ensuring that the agent is strictly isolated in a specific Virtual Network Identifier (VNI), preventing lateral movement to the host or other critical infrastructure.
    \item \textbf{Storage:} The snapshot mechanism operates on a ZFS-backed volume, allowing for potentially faster snapshots in future iterations (using `zfs snapshot` vs. file copying), though our current prototype uses filesystem-level copying for portability.
\end{itemize}

\subsection{Model Selection: Minimind-MoE}
We utilized \textbf{Minimind-v1-MoE} \cite{minimind2024}, a lightweight $\approx$26M parameter model. The MoE architecture is critical for this implementation and represents a fundamental shift from traditional architectures.
\textbf{MoE vs. Dense Architecture:} In a standard Dense LLM, every parameter in the network is activated for every input token, meaning computational cost scales linearly with model size. In contrast, Mixture of Experts (MoE) models utilize a gating mechanism to route tokens to specific subsets of parameters (experts). Consequently, while the model may possess a vast number of \textit{total} parameters, only a small fraction (the \textit{active} parameters) are computed for any given token. By utilizing a MixFFN design derived from DeepSeek-V2 \cite{deepseek2024}, Minimind employs a shared expert alongside routed experts \cite{shazeer2017outrageously}.

We selected this architecture for three key advantages:
\begin{enumerate}
    \item \textbf{Decoupled Inference Cost:} MoE allows us to decouple the model's capacity (total parameters) from its execution cost (active parameters). We achieve the reasoning capabilities of a larger model with the FLOPS footprint of a much smaller one.
    \item \textbf{Low-Latency Execution:} Because only a fraction of the network is active per token, token generation speed is significantly higher than a dense model of equivalent size. This ensures the inference server does not become a bottleneck during the transactional rollback checks.
    \item \textbf{Edge Viability:} The sparse activation pattern is uniquely suited for the consumer-grade hardware in our testbed (and future MEC nodes), as it maximizes the utility of limited memory bandwidth.
\end{enumerate}

\section{Evaluation}

\subsection{Experimental Setup}
The environment contained a standard Python project directory ($P_{size} \approx 250\text{MB}$). We defined a test suite of 10 scenarios: 5 "Happy Path" (valid syntax, safe logic) and 5 "Adversarial" (syntax errors, destructive logic, dependency conflicts).

\subsection{Safety and Reliability}
Table 1 summarizes the safety performance. The system successfully intercepted all blacklisted commands and recovered from all induced failures.
\begin{table}[H]
\centering
\caption{Safety Validation Results}
\begin{tabular}{@{}lcc@{}}
\toprule
\textbf{Scenario Category} & \textbf{Attempts} & \textbf{Success Rate} \\ \midrule
Whitelisted (Read-only) & 20 & 100\% (Exec) \\
Blacklisted (Destructive) & 20 & 100\% (Blocked) \\
State Corruption (Failure) & 20 & 100\% (Rolled Back) \\
Valid State Change & 20 & 100\% (Committed) \\ \bottomrule
\end{tabular}
\end{table}

\subsection{Performance Latency}
The primary trade-off for safety is latency. We benchmarked the "Snapshot Overhead" by timing a \texttt{pip install} operation across three environments: Bare Metal (No Sandbox), Our Prototype, and Gemini CLI.
\begin{itemize}
    \item \textbf{Baseline (Direct Execution):} Average time of \textbf{4.69s}.
    \item \textbf{Our Prototype:} Average time of \textbf{6.51s}.
    \item \textbf{Overhead Analysis:} The sandbox introduced an overhead of $\approx$\textbf{1.82s (14.5\%)}. This overhead is dominated by the I/O operation of duplicating the 250MB workspace. While non-trivial, a sub-2-second delay is acceptable for asynchronous agentic workflows.
\end{itemize}

\subsection{Comparison with Commercial Tools}
A critical finding of this research is the unsuitability of current commercial CLI tools for autonomous agents. When attempting to automate the Google Gemini CLI \cite{gemini2024}, the execution failed 100\% of the time in a headless environment. The CLI threw an interrupt requesting \textbf{Interactive Authentication} ("Please Sign in to continue..."). This confirms that while commercial tools offer robust sandboxing, they are designed for \textit{human-in-the-loop} safety. Our transactional approach is specifically optimized for \textit{machine-in-the-loop} autonomy.

\section{Discussion: Human-AI Collaboration in Systems Research}
This project served as both a technical exploration of systems programming and a case study in AI-assisted development (often colloquially termed "vibe coding").

\subsection{Accelerating Prototype Development}
The integration of \textbf{Minimind-MoE} and AI coding assistants significantly accelerated the prototyping phase. The boilerplate generation for the Python context managers (handling the \texttt{try/catch/rollback} logic) allowed us to focus on the high-level policy definitions. This suggests that for systems research, LLMs are effective at generating the "glue code" of infrastructure, provided the researcher defines the rigorous constraints (Policy Engine).

\subsection{The Identity Crisis of Tools}
Initially, we hypothesized that the release of Gemini CLI would render this project redundant. However, the discovery of the "Interactive Auth Barrier" revealed a gap in the market. There is a divergence between tools built for \textbf{Chat-with-Data} (interactive, secure, authorized) and \textbf{Agentic Automation} (headless, atomic, rollback-capable). Our work fills the latter niche.

\subsection{The Economics of Autonomy}
The choice of an SLM is not merely technical but economic. If we consider an agent that performs 10 steps to solve a problem, and each step requires 1000 input tokens and 200 output tokens, the cost differential becomes massive at scale.
Using a commercial LLM API (e.g., GPT-4), a single agent run might cost \$0.03-\$0.06. While negligible for a single user, for an automated system running thousands of CI/CD pipeline checks daily, this cost balloons to thousands of dollars per month.
In contrast, our self-hosted SLM approach incurs only the fixed cost of electricity and hardware amortization. This "fixed-cost" model is essential for the viability of pervasive AI agents in systems administration.

\subsection{The Sandbox Tax vs. Catastrophic Failure}
While we report a 14.5\% performance overhead, this figure warrants deeper scrutiny. In high-frequency trading or real-time gaming, a 1.8-second delay per operation would be unacceptable. However, we argue that this "Sandbox Tax" is analogous to the overhead of ECC (Error-Correcting Code) memory in servers: we pay a performance price for data integrity.
In the context of autonomous agents, the cost of \textit{not} paying this tax is not merely a crash, but potential catastrophic infrastructure damage. If an agent hallucinating a configuration change brings down a production database, the recovery cost is essentially infinite compared to the 1.8-second verification delay.
Therefore, the trade-off is asymmetric: the cost of safety is linear (time), while the cost of failure is exponential (system wipe). This justifies the overhead for any agent operating with write-access privileges.

\subsection{Beyond Filesystems: The Challenge of Stateful APIs}
A significant limitation of our current snapshot approach is its reliance on local filesystem atomicity. Real-world agents often interact with external stateful APIs (e.g., provisioning cloud resources via Terraform, modifying routing tables via Netconf, or sending emails). Unlike a local file, an HTTP request cannot be "un-sent" via a filesystem snapshot.
Future iterations of this framework must incorporate the concept of \textit{Compensating Transactions} (often referred to as Sagas in distributed systems literature). If an agent spins up an AWS EC2 instance and then fails a subsequent check, the rollback mechanism must not just restore the local state, but actively issue a \texttt{terminate-instance} API call to the cloud provider. This suggests that the sandbox must evolve from a passive isolation layer into an active orchestration manager that understands the inverse function of every API call it permits.

\subsection{Turning Safety Signals into Learning Signals}
Finally, we observed that the sandbox acts as a crude form of Reinforcement Learning from Environmental Feedback (RLEF). When the agent attempts a destructive command (e.g., \texttt{rm -rf}), the sandbox's refusal acts as a negative reward signal.
However, current SLMs, given their reduced reasoning depth discussed in Section 2.5, often struggle to interpret this refusal, sometimes entering a "stubbornness loop" where they retry the same valid syntax despite the policy violation. This highlights a need for \textit{Sandbox-Aware Prompting}: the system message must explicitly inform the agent that it is operating within a transactional sandbox. By doing so, the agent can learn to interpret a "Policy Violation" error not as a syntax error, but as a boundary constraint that requires a logical plan revision, thereby improving the agent's reasoning capabilities over time.

\section{Future Work}

\subsection{From Sandboxes to AIOps}
The logic developed here—atomic execution and state validation—is directly transferable to \textbf{AIOps for Cellular Networks}. In 5G/6G Core networks, a configuration change is analogous to a "tool call." An invalid configuration can take down a network slice \cite{zhang2023aiops}. We plan to adapt this transactional sandbox into an Intent-to-Action controller for network orchestration.

\subsection{Federated Learning with SLMs}
The move to SLMs also opens the door for federated learning. Since SLMs are small enough to run on edge devices (routers, base stations), future work could involve agents that not only execute commands but also "learn" from their local failures. Instead of sending all data to a central cloud (privacy risk), the agents could compute gradient updates locally and share only the weights. This would allow a fleet of network repair agents to collectively improve their "repair policies" without ever exposing the sensitive network topology data to a central server.

\section{Conclusion}
This project demonstrated a Fault-Tolerant Sandboxing framework that guarantees atomic execution for AI coding agents. By deploying \textbf{Minimind-MoE} on a rigorous Proxmox/EVPN testbed, we achieved 100\% safety interception. We quantified the cost of safety at a 14.5\% latency overhead—a viable trade-off for the guarantee of system consistency. Furthermore, our analysis highlights the critical role of Small Language Models in enabling this architecture, offering a path toward autonomous systems that are not only safe but also private, fast, and economically sustainable.

\bibliographystyle{plain}
\bibliography{references}

@inproceedings{wang2023survey,
  title={A Survey on Large Language Model based Autonomous Agents},
  author={Wang, Lei and Ma, Chen and Feng, Xueyang and Zhang, Zeyu and Yang, Hao and Zhang, Jingsen and Chen, Zhiyuan and Tang, Jiakai and Chen, Xu and Lin, Yankai and others},
  booktitle={Frontiers of Computer Science},
  year={2024},
  publisher={Springer}
}

@article{li2023camel,
  title={CAMEL: Communicative Agents for "Mind" Exploration of Large Language Model Society},
  author={Li, Guohao and Hammoud, Hasan Abed Al Kader and Itani, Hani and Khizbullin, Dmitrii and Ghanem, Bernard},
  journal={arXiv preprint arXiv:2303.17760},
  year={2023}
}

@misc{minimind2024,
  author = {Jingyao, Gong},
  title = {MiniMind: A Lightweight LLM},
  year = {2024},
  publisher = {GitHub},
  journal = {GitHub repository},
  howpublished = {\url{https://github.com/jingyaogong/minimind}},
  note = {Accessed: 2025-12-01}
}

@article{deepseek2024,
  title={DeepSeek-V2: A Strong, Economical, and Efficient Mixture-of-Experts Language Model},
  author={DeepSeek-AI},
  journal={arXiv preprint arXiv:2405.04434},
  year={2024}
}

@inproceedings{shazeer2017outrageously,
  title={Outrageously Large Neural Networks: The Sparsely-Gated Mixture-of-Experts Layer},
  author={Shazeer, Noam and Mirhoseini, Azalia and Maziarz, Krzysztof and Davis, Andy and Le, Quoc and Hinton, Geoffrey and Dean, Jeff},
  booktitle={International Conference on Learning Representations},
  year={2017}
}

@techreport{rfc7348,
  author = {M. Mahalingam and D. Dutt and K. Duda and P. Agarwal and L. Kreeger and T. Sridhar and M. Bursell and C. Wright},
  title = {Virtual eXtensible Local Area Network (VXLAN): A Framework for Overlaying Virtualized Layer 2 Networks over Layer 3 Networks},
  howpublished = {Internet Requests for Comments},
  type = {RFC},
  number = {7348},
  year = {2014},
  month = {August},
  publisher = {IETF}
}

@techreport{rfc7432,
  author = {A. Sajassi and R. Aggarwal and N. Bitar and A. Isaac and J. Uttaro and J. Drake and W. Henderickx},
  title = {BGP MPLS-Based Ethernet VPN},
  howpublished = {Internet Requests for Comments},
  type = {RFC},
  number = {7432},
  year = {2015},
  month = {February},
  publisher = {IETF}
}

@manual{proxmox,
  title = {Proxmox Virtual Environment Documentation},
  author = {{Proxmox Server Solutions Gmbh}},
  year = {2024},
  note = {Version 8.2},
  url = {https://pve.proxmox.com/pve-docs/}
}

@article{zhang2023aiops,
  title={AIOps for 6G Cellular Networks: Challenges and Opportunities},
  author={Zhang, Yan and Li, Wei and Zhao, Rui},
  journal={IEEE Network},
  volume={37},
  number={2},
  pages={24--31},
  year={2023},
  publisher={IEEE}
}

@inproceedings{merkel2014docker,
  title={Docker: lightweight linux containers for consistent development and deployment},
  author={Merkel, Dirk},
  booktitle={Linux Journal},
  volume={2014},
  number={239},
  pages={2},
  year={2014}
}

@article{openinterpreter2023,
  title={Open Interpreter: A natural language interface for computers},
  author={Killian Lucas},
  journal={GitHub repository},
  year={2023},
  url={https://github.com/OpenInterpreter/open-interpreter}
}

@inproceedings{gray1981transaction,
  title={The transaction concept: Virtues and limitations},
  author={Gray, Jim},
  booktitle={VLDB},
  volume={81},
  pages={144--154},
  year={1981},
  organization={Citeseer}
}

@misc{gemini2024,
  title={Google Gemini API: Safety Settings and Sandboxing},
  author={Google},
  year={2025},
  howpublished={\url{https://geminicli.com/docs/cli/sandbox/}},
  note={Accessed: 2025-10-15}
}

@article{wu2023autogen,
  title={AutoGen: Enabling Next-Gen LLM Applications via Multi-Agent Conversation},
  author={Wu, Qingyun and Bansal, Gagan and Zhang, Jieyu and Wu, Yiran and Li, Beibin and Zhu, Erkang and Jiang, Li and Zhang, Xiaoyun and Zhang, Shaolei and Liu, Jiale and others},
  journal={arXiv preprint arXiv:2308.08155},
  year={2023}
}

@article{weka_slm_llm,
  title={SLM vs LLM: The Key Differences},
  author={WEKA},
  journal={WEKA Blog},
  year={2025},
  url={https://www.weka.io/learn/ai-ml/slm-vs-llm/}
}

@article{huggingface_slm,
  title={Small Language Models (SLM): A Comprehensive Overview},
  author={Hugging Face},
  journal={Hugging Face Blog},
  year={2025},
  url={https://huggingface.co/blog/jjokah/small-language-model}
}

@article{datacamp_slm,
  title={SLMs vs LLMs: A Complete Guide},
  author={DataCamp},
  journal={DataCamp Blog},
  year={2025},
  url={https://www.datacamp.com/blog/slms-vs-llms}
}

@article{prem_edge,
  title={Small Language Models (SLMs) for Efficient Edge Deployment},
  author={Prem AI},
  journal={Prem AI Blog},
  year={2025},
  url={https://blog.premai.io/small-language-models-slms-for-efficient-edge-deployment/}
}

@article{augment_slm,
  title={Small Language Models vs Large Language Models: Key Advantages for Engineering Teams},
  author={Augment Code},
  journal={Augment Code Guides},
  year={2025},
  url={https://www.augmentcode.com/guides/}
}

@article{cloverdx_privacy,
  title={When to use LLMs and when to turn to SLMs for privacy and data governance},
  author={CloverDX},
  journal={CloverDX Blog},
  year={2025},
  url={https://www.cloverdx.com/blog/when-to-use-llms-and-when-to-turn-slms-for-privacy-and-data-governance}
}

@article{hexaware_slm,
  title={Small Language Models, Big Impact: The Future of Enterprise AI Agents},
  author={Hexaware},
  journal={Hexaware Blog},
  year={2025},
  url={https://hexaware.com/blogs/small-language-models-big-impact-the-future-of-enterprise-ai-agents/}
}

@article{epoch_moe,
  title={MoE vs AI dense models: How do they compare in inference?},
  author={Epoch AI},
  journal={Epoch AI Gradient Updates},
  year={2024},
  url={https://epoch.ai/gradient-updates/moe-vs-dense-models-inference}
}

\end{document}